# A novel method for Causal Structure Discovery from EHR data:

A demonstration on type-2 diabetes mellitus

Xinpeng Shen[a], Sisi Ma[a,b], Prashanthi Vemuri[c], M. Regina Castro[d], Pedro J. Caraballo[e], Gyorgy J. Simon[a,b]

[a]Institute for Health Informatics, University of Minnesota, Minneapolis, MN, USA

[b]Department of Medicine, University of Minnesota, Minneapolis, MN, USA

[c]Department of Radiology, Mayo Clinic, Rochester, MN, USA

[d]Division of Endocrinology, Mayo Clinic, Rochester, MN, USA

[e]Department of Internal Medicine, and Department of Health Sciences Research, Mayo Clinic, Rochester, MN, USA

Corresponding author E-mail address: simo0342@umn.edu (Gyorgy J. Simon, PhD).

**ABSTRACT**

**Introduction:** The discovery of causal mechanisms underlying diseases enables better diagnosis, prognosis and treatment selection. Clinical trials have been the gold standard for determining causality, but they are resource intensive, sometimes infeasible or unethical. Electronic Health Records (EHR) contain a wealth of real-world data that holds promise for the discovery of disease mechanisms, yet the existing causal structure discovery (CSD) methods fall short on leveraging them due to the special characteristics of the EHR data. We propose a new data transformation method and a novel CSD algorithm to overcome the challenges posed by these characteristics.

**Materials and methods:** We demonstrated the proposed methods on an application to type-2 diabetes mellitus. We used a large EHR data set from Mayo Clinic to internally evaluate the proposed transformation and CSD methods and used another large data set from an independent health system, Fairview Health Services, as external validation. We compared the performance of our proposed method to Fast Greedy Equivalence Search (FGES), a state-of-the-art CSD method in terms of correctness, stability and completeness. We tested the generalizability of the proposed algorithm through external validation.

**Results and conclusions:** The proposed method improved over the existing methods by successfully incorporating study design considerations, was robust in face of unreliable EHR timestamps and inferred causal effect directions more correctly and reliably. The proposed data transformation successfully improved the clinical correctness of the discovered graph and the consistency of edge orientation across bootstrap samples. It resulted in superior accuracy, stability, and completeness.

1. Introduction

Discovering causal relationships is fundamental to medical sciences[1]. A relationship is causal if manipulating one variable changes the other variable[2]. Causal relationships are in contrast with associative relationships, where manipulating (improving) one disease does not necessarily bring along improvement in another. If diseases are related to each other causally, successfully treating one will improve the others. Studying interventions in medical sciences is thus predicated on understanding the underlying causal relationships.

While clinical trials have been the gold standard for determining causality, they are resource intensive, sometimes infeasible or unethical. An alternative to clinical trials is causal structure discovery (CSD), where the goal is to discover all causal relationships among variables, known as the causal structure, from observational data through computational algorithms. These algorithms, in general, require large training data sets especially when the number of variables is large. Through the wide-spread adoption of electronic health record (EHR) systems, large quantities of data have been accumulated, which can serve as the data source for a wide range of studies[3]. However, currently existing CSD methods fall short on EHR data in the following ways.

*Study design considerations.* EHR data as it exists in the system serves the purpose of documenting care and does not follow any study design. While patients may be physically present in a health care facility, without performing the appropriate diagnostic tests, many diseases are not diagnosable. Such patients are not under observation for those diseases. Another example of study design consideration is a patient with a pre-existing disease. This patient is not at risk of developing the first incidence of that disease and the subsequent diagnoses that the patient continues to receive simply indicate ongoing treatment as opposed to new incidences of the disease. Failure to account for such examples can make a relationship appear stronger or weaker and can even reverse the effect direction. None of the currently existing CSD algorithms makes these study design considerations.

*Unreliable diagnosis time stamps.* The time stamp of a diagnosis often does **not** coincide with the onset time of the disease, but rather reflects the documentation time. In some cases, the temporal ordering of diseases may be reversed. Partly for this reason, CSD algorithms occasionally report "causal" relationships that are in the opposite direction of the natural disease progression.

*Unoriented edges.* Even when a clear causal direction exists and is not masked by data artifacts, CSD algorithms can have difficulty identifying the correct effect direction. For example, if there are only two associated variables A and B, most causal discovery algorithms will not be able to infer the direction between them without additional information, because both possible directions, A → B and B → A, yield identical likelihood and independence test results. In such cases, currently existing methods report an unoriented (directionless) relationship.

We apply our methodology to a large longitudinal EHR data set using type-2 diabetes mellitus (T2DM) as a demonstration. T2DM is an ideal demonstration, because its comorbidities and complications are interrelated and their causal relationships are mostly known. T2DM is also clinically significant, representing a growing epidemic that affects 30.3 million people in the United States (9.4% of the population)[4]. It is a chronic disease with many potential causes and significant medical complications, including stoke, neuropathy, cardiovascular and peripheral vascular disease. Historically, there have been debates over the causal relationships related to diabetes[5], like the effect of hyperlipidemia or statin use on T2DM[6-8]. Studies have shown that these relationships are often confounded with other disease conditions[9].

Our objective is to develop methodologies that address EHR-related challenges. Specifically, (1) we propose a data transformation procedure that distinguishes new incidences from pre-existing conditions, allowing the subsequent CSD algorithm to make the appropriate study design considerations. This leads to an operational definition of causality which is closer to clinical interpretation even when used in conjunction with existing CSD algorithms. (2) We also develop a CSD algorithm based on the (Fast) Greedy Equivalence Search (FGES) algorithm[10, 11] that can infer the direction of causal relationships more robustly using longitudinal information and takes the above study design considerations into account. While we use T2DM as our use case, we expect our methods to generalize to other diseases, typically chronic diseases, that exhibit similar characteristics and suffer from the same EHR shortcomings.

We apply our methodology to a large longitudinal EHR data set of 57,332 patients from Mayo Clinic (MC). We compare our results to those obtained from the FGES algorithm both on the original data, and also from the data set that is transformed using the proposed data transformation method. We then evaluate the discovered causal structure internally and validate it externally on an independent data set from Fairview Health Services (FHS) with 79,486 patients.

## 2. Related Works

Learning biomedical knowledge from EHR data continues to receive considerable research attention. The high-dimensional, biased, irregular temporal, and sparse nature of EHR data creates challenges for extracting accurate, actionable (causal), and in some cases even interpretable biomedical relationships.

The focus of this manuscript is on causal inference and particularly on causal structure discovery. A causal effect between variables A and B has the interpretation that A affects B, as opposed to merely being associated with B, or being predictive of B. Causal inference refers to the problem of estimating the size of this effect, while causal structure discovery refers to the problem of extracting all causal relationships among pairs of variables (without necessarily estimating the size of these effects). Traditional algorithmic causal discovery methods typically fall into three categories: constraint-based (e.g. PC[12], FCI[13, 14]), score-based (e.g. FGES), and hybrid methods (e.g. MMHC[15]). Many of the earlier methods have strong assumptions regarding the structure of the data generating process. Recent methods extend the previous methods by relaxing many of the assumptions: handling non-linear casual relationships[16], cyclic causal relationships[17], individualized causal structures[18, 19]. Despite the advances in methods for causal structure discovery, to our knowledge, no method can overcome the specific challenges posed by the EHR data. Our proposed method is inspired by the score-based methods and designed to work effectively with EHR data.

Another related but distinct area of work is causal inference, or the estimation of causal effect sizes. Traditionally, causal effect size estimation methods assume that the causal structure is known and estimate causal effect sizes from models of the outcome (directly adjusting for confounders[20]) or by de-confounding the data through propensity models[21]. The flexibility afforded by deep learning models opens new possibilities for causal inference. For example, Bahadori et al.[22] proposed a new causal regularization which outperforms traditional L1 norm on causal accuracy. Louizos et al.[23] designed a causal effect variational autoencoder based on variational autoencoder (VAE) to incorporate latent space modeling. All the above newly proposed methodologies were designed to incorporate two important causal concepts, latent variables and confounding effects, into the model architecture. As a result, these models offer more causal interpretability to its results compared to traditional deep learning models. Often, the question that motivates causal discovery, aims to make the causal relationships among variables explicit, rather than estimating causal effect sizes. Deep learning models, without explicit interpretability of the encoded relationships, only offer effect size estimates with possible causal interpretation. Our goal is to make the causal structure among variables explicit rather than to estimate effect sizes.

We note that deriving models for causal relationship is related to but fundamentally different from model interpretation. There is considerable effort directed at addressing the interpretability of black-box models. Recently, Kwon et al.[24] proposed a visualization toolbox to increase interpretability and interactivity of RNNs. Users can use the toolbox to investigate patient history, view patient's contribution, and conduct what-if analysis, and input domain knowledge into the training process. Liang et al.[25] proposed Model-agnostic Effective Efficient Direct (MEED) Instance-wise Feature Selection (IFS) to evaluate the degree to which the selected feature explains the conditional distribution of the target. The model uses the result of the prediction task to rule out unimportant features. While these methods offer better interpretability over traditional black box machine learning models, these methods neither make relationships among variables explicit nor do they afford causal interpretation to results.

### 3. Material and methods
3.1 Data

This retrospective cohort study utilized two separate large EHR data sets from two independent health systems, Mayo Clinic and Fairview Health Services, with 57,332 and

79,486 patients, respectively. Two cross sections (2002-2005 and 2005-2008 for MC; and 2007-2010 and 2010-2013 for FHS) were defined. Cross sections differed between MC and FHS due to data availability. We extracted diagnoses, prescriptions, laboratory results, and vital signs from the two EHR data sets with the same inclusion and exclusion criteria: patients must have at least two blood pressure measurements, one before the first cross section and one after the second cross section; age 18+ at the end of the first cross section, and sex and age must be known.

Diagnosis codes were aggregated into meaningful disease categories (e.g. Type-2 diabetes mellitus) and medications were rolled up into NDF-RT therapeutic subclasses. Lab results were categorized based on cutoffs from the ADA guidelines[26]. The variables are shown in Table 2 as percentage of patients with the given treatment, diagnosis or abnormal lab result for both the MC and the FHS data sets. Chi-square test was conducted to determine whether the two datasets differ for each variable. (except for age, which is a continues variable, a t-test was conduct)

### 3.2 Methods

Although numerous CSD algorithms exist[27, 28], we reviewed FGES because: (1) it is one of the most popular CSD algorithms; (2) it has been shown to produce excellent results[29, 30]; (3) it is closely related to the proposed algorithm; and (4) we used it as the baseline algorithm in our evaluation. FGES evaluates a causal structure by assigning a goodness of fit score. It optimizes this score by starting with an empty structure without any edges, and adding edges iteratively until further adding another edge no longer improves the fit. Then, FGES tries to remove edges one at a time as long as the fit improves. The output of FGES is a graphical representation of the causal structure, called a pattern[12]. Like a directed acyclic graph (DAG), a pattern consists of nodes and edges, but unlike a DAG, where every edge has a direction, edges in a pattern can be undirected. Edges are undirected when the scores corresponding to the different orientations of the edge are the same. From a different point of view, a pattern represents a set of DAGs satisfying the same independence criteria. Theoretically, FGES has been shown to be consistent under a set of assumptions, including no unobserved confounders, and infinite sample size[11, 31]. We used the FGES implementation in R software, package 'rcausal', version 1.1.1 with a penalization parameter of 2.

#### 3.2.1 The proposed method

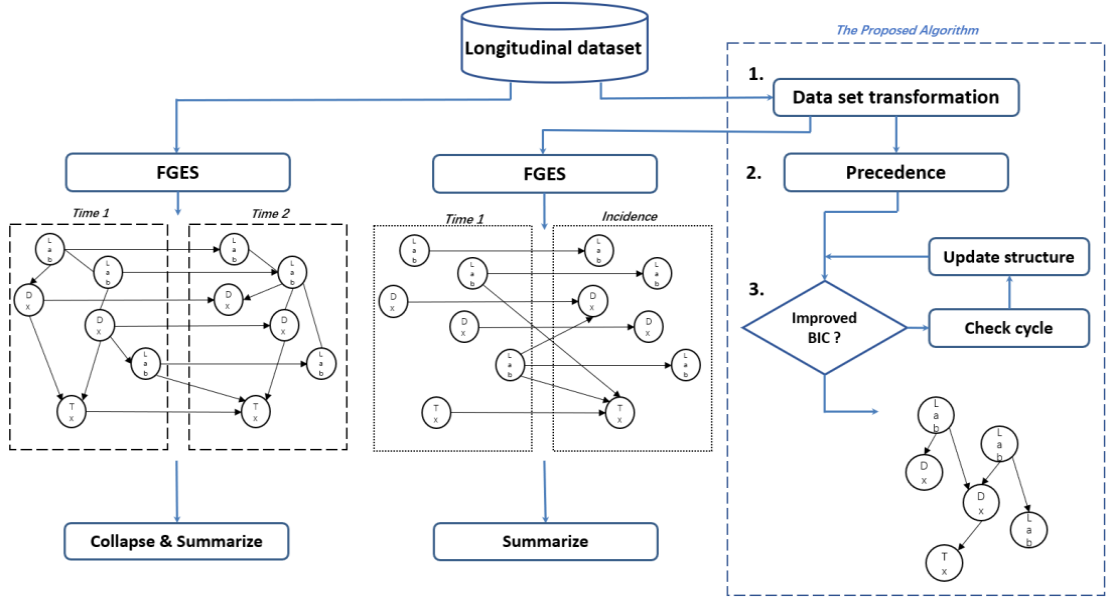

Figure 1. The proposed methods and the overview of the internal evaluation.

### 3.2.2 Data set transformation

**Condition** is defined as a diagnosis, a prescription, an abnormal lab result or vital sign. A condition is **incident** if it is first developed in the current cross section, the patient was under observation in the previous cross sections and the patient did not have signs of the condition before the current cross section. Conversely, a condition is **pre-existing** if the patient had it in any previous cross section. Since we considered progressive chronic conditions, the pre-existing condition implies that the patient is no longer at risk of developing the condition in the future.

The proposed data set transformation method utilizes the above definitions and transforms the data in the current cross section to differentiate between incidence and pre-existing conditions.

### 3.2.3 Precedence

Consider patients who already developed both conditions A and B in cross section 2. The proportion of patients who newly developed B but had pre-existing A is

$$P(A^{(1)}, \bar{B}^{(1)} | A^{(2)}, B^{(2)}), \qquad (1)$$

where the superscript represents the cross section, and the bar on the variable represents not having the condition. Conversely, the proportion of patients who developed A after B is

$$P(B^{(1)}, \bar{A}^{(1)} | A^{(2)}, B^{(2)}). \qquad (2)$$

Let $\mathcal{V} = \{V_1, V_2, \ldots, V_m\}$ denote the set of $m$ conditions. For a pair of conditions $V_i$ and $V_j$ in cross sections 1 and 2, $V_i \prec V_j$ (pron. "$V_i$ **precedes** $V_j$") if and only if:

$$\frac{P\left(V_i^{(1)}, \bar{V}_j^{(1)} | V_i^{(2)}, V_j^{(2)}\right)}{P\left(V_j^{(1)}, \bar{V}_i^{(1)} | V_i^{(2)}, V_j^{(2)}\right)} > w; \qquad (3)$$

where $w$ is called the precedence ratio. Note that precedence implies neither causation nor association; however, if a causal effect exists, it must follow the precedence direction. This requirement follows from the assumption that causal effects do not progress backwards in time.

We construct a set $\mathcal{C}$ of all pairs $(V_i, V_j)$, where $V_i \prec V_j$ and $i, j = 1, \ldots, m, i \neq j$. Note that the precedence ratio $w$ controls the "false positive" rates. It can be determined such that the proportions are statistical-significantly different, can be optimized by cross-validation, or can be chosen arbitrarily.

### 3.2.4    Search and structure update

Given the set $\mathcal{C}$, we construct the causal graph $\mathcal{G}$ by iteratively adding edge $(V_i, V_j)$ from $\mathcal{C}$ that maximizes the goodness of fit of $\mathcal{G}$. The orientation of this edge must be consistent with the precedence relationship between $V_i$ and $V_j$.

The goodness of fit is defined by the BIC criteria. Let $X^{(1)}, X^{(2)}$ denote the data sets collected at the two distinct cross sections, where $X^{(2)}$ follows $X^{(1)}$. The likelihood of the $\mathcal{G}$ is

$$\begin{aligned}
\mathcal{L}(\mathcal{G}|X^{(1)}, X^{(2)}) &= P(X^{(2)}|X^{(1)}, \mathcal{G}) \\
&= \prod_i \prod_{v \in \mathcal{V}} P\left(v_i^{(2)} \big| x_i^{(1)}, \mathcal{G}\right) \\
&= \prod_i \prod_{v \in \mathcal{V}} P\left(v_i^{(2)} \big| pa(v, \mathcal{G})_i^{(1)}\right),
\end{aligned} \quad (4)$$

where $x_i^{(t)}$ is the observation vector for subject *i* at the cross section *t*; $v_i^{(t)}$ is the observation of condition *v* for subject *i* at the cross-section *t*; and $pa(v, \mathcal{G})_i^{(1)}$ is the observation vector for the parents of (have direct causal effect on) variable *v* in the causal structure $\mathcal{G}$, at cross section 1 for subject *i*.

The algorithm estimates $P\left(v_i^{(2)} \big| pa(v, \mathcal{G})_i^{(1)}\right)$ using logistic regression on the subjects who do not have *v* at the first cross section and are under observations for both cross sections. Finally, the BIC score is

$$\mathrm{BIC}(\mathcal{G}) = -2\ln\mathcal{L}(\mathcal{G}|X^{(1)}, X^{(2)}) + \ln(n)|\mathcal{G}|, \quad (5)$$

where n is the number of observations that are common in the two cross sections, and $|\mathcal{G}|$ is the number of edges in the causal structure $\mathcal{G}$.

**Algorithm 1:** The proposed causal search algorithm.

**Input:** the precedence set $\mathcal{C}$; the data set $X$
**Output:** the discovered DAG, $\mathcal{G}$

1. Initialize $\mathcal{G} = \emptyset$
2. **while** the candidate set $\mathcal{C}$ is not empty and $\Delta BIC > 0$ **do**
3.     **foreach** candidate pair $(V_i \prec V_j) \in \mathcal{C}$ **do**
4.         $\Delta BIC = BIC(X, \mathcal{G}) - BIC\left(X, \mathcal{G} \cup (V_i \prec V_j)\right)$
5.         **if** $\Delta BIC > 0$ **then**
6.             $\mathcal{G} = \mathcal{G} \cup (V_i \prec V_j)$ // add the candidate pair on the current graph
7.             $\mathcal{C} = \mathcal{C} \setminus (V_i \prec V_j)$ // Remove the candidate pair from $\mathcal{C}$ else stop
8.         **end**
9.     **end**
10. **end**
11. Return $\mathcal{G}$

Algorithm 1. The proposed causal search algorithm.

Algorithm 1 describes the proposed algorithm for constructing the causal graph $\mathcal{G}$. $\mathcal{G}$ is a DAG, with nodes representing conditions and edges representing causal effects between a pre-existing and an incident condition.

### 3.2.5 Interpretation

Consider two nodes A and B representing two conditions, where A is pre-existing and B is incident. A causes B (A → B) if and only if:
   a) A precedes B (as defined before)
   b) A is a predictor of B (A and B are associated)
   c) If there is a pre-existing condition C, such that C → A and C → B, then A and B are causally related even after adjusting for C
   d) B → A does not satisfy at least one of the above three conditions.

Note that the typical definition of causation does not imply a and d.

## 3.3 Experiment design

### 3.3.1 The internal comparison

Figure 1 provides an overview of the internal comparison. To isolate the effects of the proposed data transformation method and the proposed search algorithm, we discovered a causal graph in three different ways. First, we ran FGES directly on the raw data; second, we ran FGES on the transformed data; and finally, we ran the proposed search algorithm on the transformed data. We referred to the resultant graphs as **FGES+raw**, **FGES+transf**, and **Proposed**, respectively. Comparing FGES+raw and FGES+transf isolated the effect of the proposed transformation method, and comparing FGES+transf and Proposed highlighted the effect of the proposed search algorithm.

### 3.3.2 The external validation
The graphs were discovered from the two EHR data sets independently, and were compared.

## 3.4 Evaluation method

Evaluating CSD algorithms using non-simulated datasets is always challenging, due to the lack of the true causal structure to compare with. The gold standard for proving a causal relationship is through a clinical trial which is often highly impractical when the task is to evaluate a causal structure rather than a single causal relationship. In T2DM, the full causal graph is unknown, however, numerous relationships have been discovered through clinical trials and additional relationships are broadly acknowledged by clinicians. Such relationships provide some "ground truths" that can be used for evaluation, but in the absence of a complete true graph, the absolute completeness of the discovered causal structures cannot be evaluated. An alternative way of verifying the discovered causal relationships is by pointing out "incorrect" relationships.

### 3.4.1 The internal comparison
We evaluated the CSD algorithms internally from three perspectives.

*Directional stability*

If an edge represents a true causal relationship, we expect the algorithm to reliably discover it in multiple bootstrap iterations. An edge is **ambiguous** if it was discovered in more than one orientation during the 1000 bootstrap iterations. Some of the ambiguous edges could be very infrequent. Thus, we further defined an edge as **frequent-ambiguous**, if the edge was discovered frequently (appears in 50% of the bootstrap iterations) and did not have a dominant orientation (both orientations appear in at least 30% of iterations when this edge is reported). For each method, we reported the proportion of ambiguous and frequent-ambiguous edges.

*Clinical correctness*

The conditions were grouped into five disease categories: obesity (Ob), hyperlipidemia (Hl), hypertension (Htn), diabetes mellitus (Dm), and complications (Compl). Table 1 describes the variables that each disease category consists.

**Table 1. Disease categories**

| Disease category | Label | Component variables |
|---|---|---|
| Obesity | Ob | `bmi.25, bmi.30, ob.dx` |
| Hyperlipidemia | Hl | `hdl, hl.dx, hl.tx, ldl, trigl` |
| Hypertension | Htn | `dbp, htn.dx, htn.tx, sbp` |
| Diabetes Mellitus | `Dm` | `dm.dx, dm.tx, fasting.100, fasting.125, predm.dx, a1c.57, a1c.65` |
| Complications | `Compl` | `cad, cevd, chf, crf, mi, stroke` |

For each algorithm, the causal structure was discovered from 1000 bootstrap iterations. Edges between the variables were aggregated into edges between their corresponding disease categories. For example, a `bmi.25` → `sbp` edge will be transformed to ob → Htn. FGES can report unoriented edges, these are ignored. We defined the **dominant direction** between two disease categories as the direction in which the majority of the reported edges were oriented. Then, for every pair of disease categories, we reported the proportion of edges pointing in the dominant direction over edges that were discovered between the pair. We called this proportion, the **confidence of orientation**. When the direction of progression between disease categories is known, we expected the dominant direction to coincide with this direction with a confidence of orientation close to 100 percentage.

*Edge-level details*

We designed five heuristics to evaluate each discovered causal structure at the edge level. Edges that do not satisfy these heuristics are considered incorrect.

*H1. Natural diseases progression*. We expect the edges to point in the direction of worsening labs/vitals or from a disease to its complications. Edges between nodes representing the same lab test at two different time points should point towards worsening lab results e.g. `fasting.100` → `fasting.125`. (Recall that `fasting.125` denotes the *first time* fasting glucose exceeded 125 mg/dL; `fasting.100` occurred earlier.) Edges between diagnoses should point from a risk factor to an overt disease or complication: e.g. `dm.dx` → `stroke`.

*H2. Diagnostic criterion.* We expect edges between a diagnosis and its defining lab or vital to point from the lab or vital to the diagnosis (e.g. `sbp` → `htn.dx`).

*H3. Treatment criterion*. We expect abnormal lab results or vital signs to trigger the treatment (e.g. `ldl` → `hl.tx`).

*H4. Indirect diagnostic criterion*. For an edge A → B, where A and B are both diagnoses, we expect that an A.lab → B edge also exists, where A.lab is the defining lab or vital sign of diagnosis A.

*H5. Indirect treatment criterion*. Consider a disease with a diagnostic lab result, a treatment, and a downstream complication. If an edge is reported from the treatment to the complication, there should also be an edge from the lab result to the complication.

For each heuristic, we reported the percentage of edges pointing in the wrong direction among the edges to which the heuristics applied.

### 3.4.2 Completeness

Without a ground truth graph, we cannot evaluate absolute completeness, only completeness relative to the graph discovered by another algorithm, FGES+transf. We chose FGES+transf because it was trained on the same transformed data as the proposed method. We examined and discussed edges that were discovered by the FGES+transf but not by the proposed algorithm. Note that the graphs compared were discovered using the complete data without bootstrap.

### 3.4.3 The external validation

For the external validation, we performed 1000 bootstrap replications on both data sets independently using the proposed method. On each data set, all edges from the 1000 graphs were pooled. We compared the two sets of pooled edges and pointed out the edges that were discordant between the structures discovered from the MC and FHS data.

## 4. Result

### 4.1 Description of the data sets

**Table 2. Characteristics of the MC (N = 57332) and FHS (N = 79486) data sets.** For continuous variables, mean (sd) is indicated; for binary variables, percentage of positive is indicated.

|  | Label | MC | FHS | P value |
|---|---|---|---|---|
| **Demographics** | | | | |
| age | age | 48.1 (18.2) | 50.4 (14.6) | 0.000 |
| male | male | 0.43 | 0.34 | 0.000 |
| **Vitals & labs** | | | | |
| BMI $\geq$ 25 & $\leq$ 30 | bmi.25 | 27.1 | 27.5 | 0.097 |
| BMI $\geq$ 30 | bmi.30 | 32.6 | 43.1 | 0.000 |
| SBP $\geq$ 140 | sbp | 10.3 | 4.5 | 0.000 |
| DBP $\geq$ 90 | dbp | 2.3 | 1.6 | 0.000 |
| LDL $\geq$ 130 | ldl | 18.4 | 15.4 | 0.000 |
| HDL abnormal | hdl | 20.2 | 24.6 | 0.000 |
| Triglyceride $\geq$ 150 | trigl | 22.6 | 17.6 | 0.000 |
| Fating plasma glucose 100-125 | fasting.100 | 24.4 | NA | NA |
| Fating plasma glucose > 125 | fasting.125 | 11.9 | NA | NA |
| A1C > 5.7 and A1C < 6.5 | a1c.57 | NA | 6.8 | NA |
| A1C $\geq$ 6.5 | a1c.65 | NA | 7.0 | NA |
| ***Diagnoses*** | | | | |
| Hypertension | `htn.dx` | 28.4 | 30.6 | 0.000 |
| Obesity | `ob.dx` | 11.5 | 11.3 | 0.320 |
| Hyperlipidemia | `hl.dx` | 31.9 | 36.4 | 0.000 |
| Pre-diabetes mellitus | `predm.dx` | 0.9 | 2.4 | 0.000 |
| Diabetes mellitus | `dm.dx` | 7.9 | 9.5 | 0.000 |
| Chronic renal failure | `crf` | 1.2 | 0.2 | 0.000 |
| Coronary heart failure | `chf` | 2.4 | 1.2 | 0.000 |
| Coronary Artery Disease | `cad` | 9.4 | 5.6 | 0.000 |
| Myocardial infarction | `mi` | 2.4 | 0.9 | 0.000 |
| Cerebrovascular disease | `cevd` | 3.6 | 1.8 | 0.000 |
| Stroke | `stroke` | 1.2 | 0.6 | 0.000 |
| ***Treatments*** | | | | |
| Hypertension | `htn.tx` | 20.6 | 31.5 | 0.000 |
| Hyperlipidemia | `hl.tx` | 15.7 | 24.6 | 0.000 |
| Diabetes mellitus | `dm.tx` | 4.4 | 7.2 | 0.000 |

Table 2 presents a description of the datasets extracted from the two EHR systems, Mayo Clinic and Fairview Health Services. The two patient populations differ in all respects except the prevalence of obesity diagnosis. Note that Mayo Clinic uses fasting plasma glucose while FHS uses HbA1c.

4.2  The internal comparison

4.2.1  Directional stability

**Table 3. Directional stability**

| | **Ambiguous** | **Frequent-ambiguous** |
|---|---|---|
| 1. FGES+raw | 79% | 45% |
| 2. FGES+transf | 38% | 24% |
| 3. Proposed | 1% | 0% |

We extracted the causal graphs from the MC data set using (1) FGES on the raw data (FGES+raw); (2) FGES on the transformed data (FGES+transf); and (3) the proposed algorithm (Proposed). Among the edges discovered by any of the algorithms, Table 3 shows the percentage of ambiguous and frequent-ambiguous edges. Approximately, 79% of edges found by FGES+raw are ambiguous. However, when the proposed data transformation was applied (FGES+transf), this number dropped to 38%, and when the proposed method was applied, it dropped further to a mere 1%. About 45% (24%, respectively) of the edges discovered by FGES+raw (and FGES+transf, respectively) were frequent-ambiguous while none of the edges discovered by the proposed method was frequent-ambiguous.

The proposed algorithm achieved the best directional stability, while FGES+raw was the least directionally stable. Note that the directional stability does not assess the correctness of the edges.

### 4.2.2 Clinical correctness

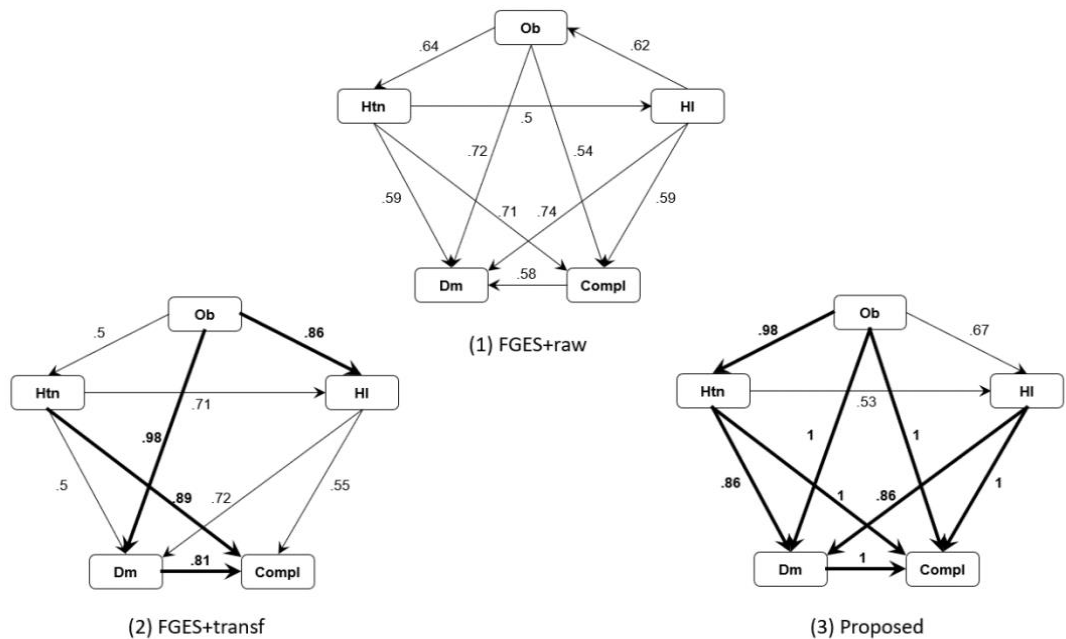

Figure 2. Clinical correctness

Figure 2 contains the DAGs obtained from the three algorithms after grouping all nodes into their corresponding disease categories (details on the grouping can be found in table 1). The edges were oriented by the dominant direction and the number next to the edge is its confidence of orientation (for definition see section 3.4.1). Edges are **bold** if the confidence of the orientation is higher than 80%.

The graphs are similar, but differing in that (i) FGES+transf did not report the 'ob' to 'compl' edge, (ii) some of the edges in FGES+raw have the incorrect orientation, and (iii) the Proposed graph have higher confidence in the orientation of the (correct) edges..

The graphs represent what we know: many patients start with obesity and proceed towards comorbidities (HTN and HL), T2DM itself, and its complications. Not all patients follow this typical progression, but many do[32]. While it is possible that an edge pointing in the direction that opposes typical progression is correct, the Hl to Ob edge or the diabetes complications to diabetes edges in the FGES+raw graph are unlikely to be correct. The FGES+transf and the Proposed graphs contain no such known incorrect edges.

Another key difference between the methods lies in their confidence of direction. Among the three algorithms the proposed algorithm yielded the DAG with the highest number of bold edges while FGES+raw contains the least (none). In fact, the only two edges in the Proposed graph that are not bold are Ob $\rightarrow$ HL (Obesity is known to cause hyperlipidemia) and HTN $\rightarrow$ HL, which is typically correct, but does not happen in all patients [18].

### 4.2.3 Edge-level details

We continued by assessing the correctness of the edges at the node level as opposed to the disease level using the defined heuristics.

**Table 4. Heuristics**

|            | H1    | H2    | H3    | H4   | H5 |
|------------|-------|-------|-------|------|----|
| FGES+raw   | 0.375 | 0.029 | 0.025 | 0    | 0  |
| FGES+transf| 0.375 | 0.032 | 0.011 | 0    | 0  |
| Proposed   | 0     | 0.013 | 0     | 0.03 | 0  |

The table 4 shows the proportion of edges that contradict each particular heuristic.

H1 is the heuristic that shows the largest difference among the algorithms: Both FGES+raw and FGES+transf discovered edges that contradict the natural progression. Specifically, these are `bmi.30` $\rightarrow$ `bmi.25`, `fasting.125` $\rightarrow$ `fasting.100`, and `dm.dx` $\rightarrow$ `predm.dx`. Recall that variables were defined to indicate worsening (not improvement), so these edges are clearly incorrect. None of the edges found by the proposed algorithm contradicted the natural progression.

H2 and 3: While FGES+raw and FGES+transf incorrectly discovered edges from the disease diagnosis or treatment to its defining lab or vital result (2.9% and 2.5% for FGES+raw, and 3.2% and 1.1% for FGES+transf), the proposed method discovered fewer such edges (1.3% and 0.0%). At first glance, it looks reasonable that a diagnosis or treatment influences its defining lab result, but the variables are defined as the first abnormal lab result, and thus should be the trigger for the diagnosis or treatment rather than its consequence. Such edges are incorrect.

H4: Only the proposed method discovered edges (0.03 in the table) that failed to satisfy the heuristics H4. We will discuss these edges in detail in the Discussion.

### 4.3 Completeness

There were 69 and 86 edges in the graph discovered by the proposed and FGES+transf, respectively. Among the 86 edges discovered by FGES+transf, 39 were also included in the proposed graph. We did not consider these 39 edges any further. Among the remaining 47 edges, 10 were in the opposite direction as in the graph discovered by the proposed algorithm. Since we already evaluated these 10 edges for correctness and found them incorrect, we did not consider these 10 edges further.

Among the remaining 37 edges that were discovered by FGES+transf but not by the proposed algorithm, (1) 14 were present in the both orientations within the graph discovered by FGES+transf. These are not causal relationships but associations. (2) 15 out of the remaining 23 edges had common parents or were indirect relationships in the proposed method's result. These edges either failed the precedence test (in which case they did not satisfy our definition of causality) or were discarded by the proposed algorithm because they did not improve the BIC sufficiently. (3) The remaining 8 were deemed incorrect by our clinical experts and are listed in Supplement I.

### 4.4 The external validation

We compared the results from applying the proposed algorithm to both the MC and FHS data sets independently. We considered all edges that were reported in at least two-thirds of the bootstrap iterations on either data set. There were 74 such edges. Approximately, 81% of these edges coincided across the two datasets, while 19% of them differed.

**Table 5. External Validation: disagreed edges between MC and FHS EHR**

| Edge | 1. MC EHR | | 2. FHS EHR | | 3. Precedence Ratio $w$ | |
|---|---|---|---|---|---|---|
| | → | ← | → | ← | MC | FHS |
| *1. Failing precedence* | | | | | | |
| hdl-trigl | 0 | 0 | 917 | 0 | 1.5 | 2.3 |
| htn.dx-crf | 885 | 0 | 1 | 0 | 3.4 | 11.8 |
| trigl-dm.dx | 1000 | 0 | 0 | 0 | 3.1 | 1.2 |
| trigl-hl.tx | 1000 | 0 | 0 | 0 | 3.4 | 1.0 |
| *2. Fasting vs A1C* | | | | | | |
| fasting.125-dm.dx | 1000 | 0 | 0 | 0 | 2.0 | 0.4 |
| `trigl-fasting.125` | 995 | 0 | 0 | 2 | 2.2 | 0.8 |
| *3. Data errors* | | | | | | |
| `chf-mi` | 0 | 40 | 676 | 0 | 0.7 | 2.5 |
| `dbp-htn.tx` | 915 | 0 | 0 | 0 | 3.5 | 0.8 |
| `hl.dx-trigl` | 0 | 0 | 876 | 0 | 0.8 | 1.8 |
| `hl.tx-cad` | 0 | 1 | 743 | 0 | 0.8 | 2.4 |
| `ldl-hl.dx` | 721 | 0 | 0 | 0 | 1.5 | 0.9 |
| `sbp-hl.tx` | 993 | 0 | 0 | 17 | 1.9 | 0.7 |
| `sbp-htn.tx` | 1000 | 0 | 0 | 291 | 2.0 | 0.5 |
| `trigl-htn.tx` | 837 | 0 | 0 | 0 | 1.9 | 0.8 |

The table 5 shows the edges that differed between the two health systems. The first column displays the two nodes that the edge connects, followed by the number of bootstrap iterations in which the edge appeared in a particular orientation on the MC data set (column 2-3) and on the FHS data set (column 4-5). The last two columns of the table show the precedence ratio on MC and FHS, as defined in the section 3.2.3.

## 5. Discussion

In this study, we proposed a new data transformation method and a new search algorithm specifically designed for EHR data, and compared it with arguably the most popular CSD algorithm, FGES, using two independent EHR data sets. We performed an internal comparative evaluation on the MC data set and an external validation on the FHS data set.

The proposed data transformation successfully improved the directional stability of the edges (percentage of frequent-ambiguous edges discovered by FGES reduced from 45% to 24%). It also helped FGES improve clinical correctness at the disease level and improved edge-level correctness.

Our search algorithm produced a graph that had the best directional stability (none of the edges appeared in both orientations while 45% of edges discovered by FGES had both orientations), was most consistent with clinical knowledge (the directional ambiguity among diseases was between HL and HTN, which could be true), and made the lowest percentage of errors in our edge evaluation.

While FGES discovered a larger number of edges than the proposed algorithm, the proposed algorithm did not miss any of the correct edges that FGES discovered. While this does not prove that the graph discovered by the proposed method is complete, it shows that it is no less complete than the graph discovered by FGES+transf.

The proposed algorithm also had a successful external validation. Among 74 edges discovered in either data set, 81% coincided across the two data sets. The majority of the differences was a result of some of the edges failing to reach statistical significance on one of the data sets.

While the graph discovered by the proposed method was largely correct, it did make some mistakes. For example, we observed causal relationships from hypertension diagnosis to other complications (H4) without an edge from the blood pressure measurements to these complications. Ideally, the diagnosis of a disease should not cause the diagnosis of another disease; it should be the lab result that drives both (e.g. `sbp` → `htn.dx` & `sbp` → `crf`, not `htn.dx` → `crf`). Due to data artifacts the algorithm failed to discover the `sbp` → `htn.dx` edge.

*Generalizability beyond diabetes.* The proposed method can also be generalized to solve CSD problems in domains other than diabetes. The key assumptions underlying the algorithms are the (1) study design consideration emphasizing the first incidence of a condition (assuming that subsequent diagnosis codes relate to ongoing treatment) and (2) that a time gap exists between cause and effect. Accordingly, when the assumptions related to the study design are violated (i.e. acute conditions that patient completely recover from) or when no time gap between cause and effect exists, this method is not applicable or offers minimal benefit over the existing algorithms. When the assumptions are met and inaccuracies typical of EHR data exist (e.g. inaccurate time stamps), the proposed method is particularly advantageous.

*Limitations*. The algorithm requires longitudinal data with at least two time-points which might not always be available. Also, the proposed transformation needs prior knowledge on variable semantics, such as SBP and DBP being measures related to HTN. Both health systems in the study are located in the Midwest. Increased heterogeneity in the population at other locations could impact the results.

## 6. Conclusions

We have demonstrated that the graph produced by the proposed transformation and search algorithm is more stable across bootstrap iterations, contained fewer errors, is not less complete than graphs discovered by existing methods and it could be validated using longitudinal EHR from an independent health system. We conclude that the proposed method is more suitable for use in EHR data.

**Acknowledgement**
Funding: This work is supported by the National Institutes of Health (NIH) grants AG056366, TR002494, and LM011972. Contents of this document are the sole responsibility of the authors and do not necessarily represent official views of the NIH.

**Conflict of interest statement**

The authors have no competing interests to declare.